\pdfoutput=1

\documentclass[11pt]{article}

\usepackage{ACL2023}

\usepackage{times}
\usepackage{latexsym}

\usepackage[T1]{fontenc}

\usepackage{amsmath}
\usepackage{graphicx}
\usepackage{multirow}
\usepackage{booktabs}
\usepackage[utf8]{inputenc}
\usepackage{color}
\usepackage{subfig}

\usepackage{microtype}

\usepackage{inconsolata}

\title{AttenWalker: Unsupervised Long-Document Question Answering via Attention-based Graph Walking}

\author{Yuxiang Nie${}^{123}$,~~Heyan Huang${}^{123}$\thanks{~~Corresponding author},~~Wei Wei${}^{4}$,~~Xian-Ling Mao${}^{123}$ \\
        ${}^{1}$School of Computer Science and Technology, Beijing Institute of Technology\\
        ${}^{2}$Beijing Engineering Research Center of High Volume Language Information Processing \\
and Cloud Computing Applications\\
${}^{3}$Beijing Institute of Technology Southeast Academy of Information Technology\\
${}^{4}$Huazhong University of Science and Technology\\
        \texttt{\{nieyx,hhy63,maoxl\}@bit.edu.cn,weiw@hust.edu.cn}}

\begin{document}
\maketitle
\begin{abstract}
Annotating long-document question answering (long-document QA) pairs is time-consuming and expensive. To alleviate the problem, it might be possible to generate long-document QA pairs via unsupervised question answering (UQA) methods. However, existing UQA tasks are based on short documents, and can hardly incorporate long-range information. To tackle the problem, we propose a new task, named unsupervised long-document question answering (ULQA), aiming to generate high-quality \textit{long-document} QA instances in an \textit{unsupervised} manner. Besides, we propose AttenWalker, a novel unsupervised method to aggregate and generate answers with long-range dependency so as to construct long-document QA pairs. Specifically, AttenWalker is composed of three modules, i.e., span collector, span linker and answer aggregator. Firstly, the span collector takes advantage of constituent parsing and reconstruction loss to select informative candidate spans for constructing answers. Secondly, by going through the attention graph of a pre-trained long-document model, potentially interrelated text spans (that might be far apart) could be linked together via an attention-walking algorithm. Thirdly, in the answer aggregator, linked spans are aggregated into the final answer via the mask-filling ability of a pre-trained model. Extensive experiments show that AttenWalker outperforms previous methods on Qasper and NarrativeQA. In addition, AttenWalker also shows strong performance in the few-shot learning setting.\footnote{We have released our codes and data in \url{https://github.com/JerrryNie/Unsupervised-Long-Document-QA}.}
\end{abstract}

\section{Introduction}

Textual question answering (QA) is the task of answering questions given textual documents as the context. Previous works can be divided into short-document QA\footnote{Usually, the term `short-document QA' is simplified as `QA' in the literature, which refers to the QA task with a short context. We emphasize `short-document' QA in this work to distinguish it with `long-document' QA.} methods \cite{DBLP:conf/iclr/SeoKFH17} and long-document QA methods \cite{nie2022capturing}.  Short-document methods approach, and even outperform humans due to the availability of large-scale short-document QA datasets \cite{DBLP:conf/emnlp/RajpurkarZLL16}. Despite that, long-document methods still lag behind humans by a large margin since annotating long-document QA datasets \cite{DBLP:conf/naacl/DasigiLBCSG21} is time-consuming and costly.

\begin{figure}[t]
\centering
\includegraphics[width=\linewidth]{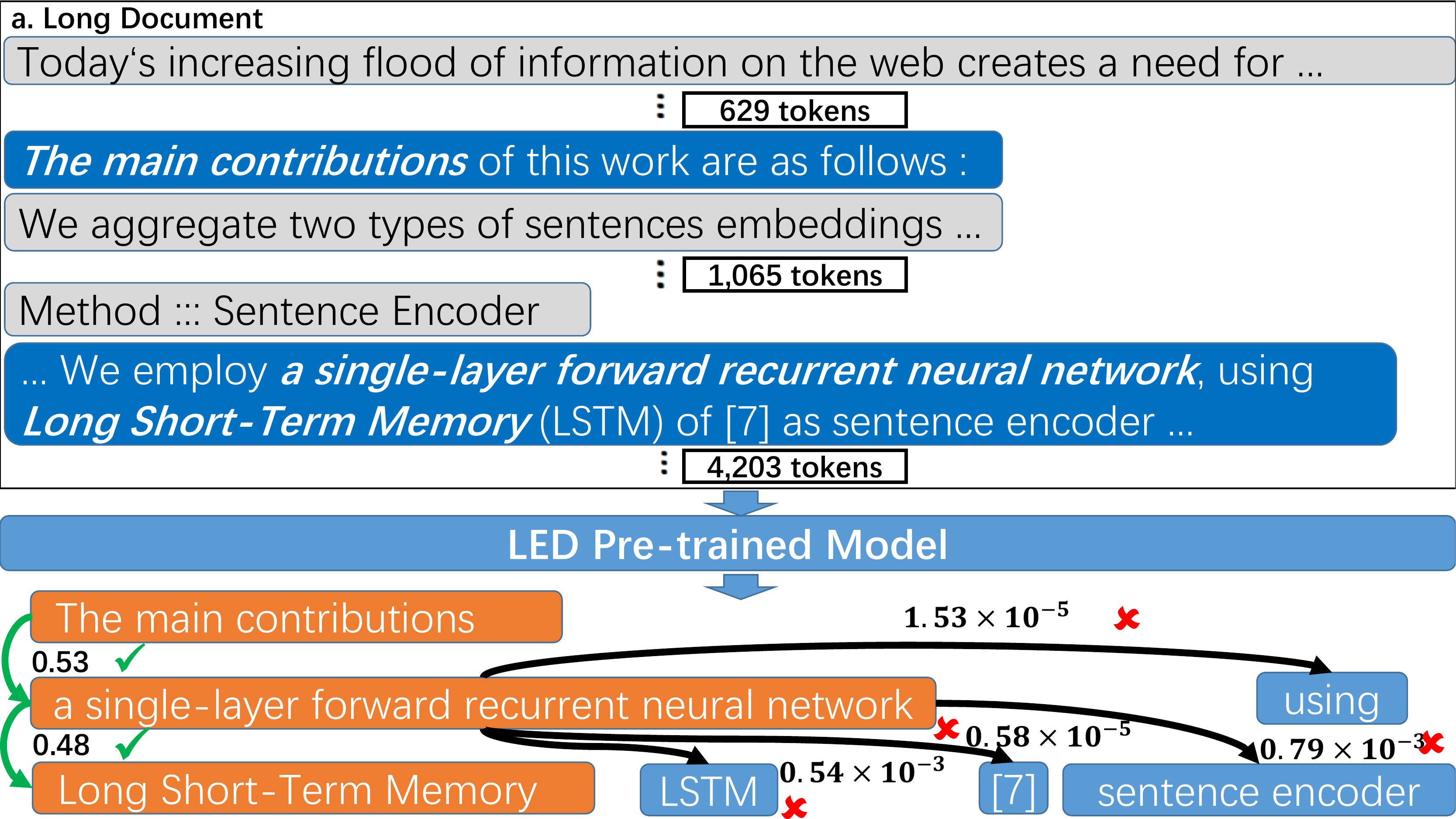}
\caption{\label{fig:illustrator}The long-range relation discovering process for a long document in Qasper dataset. The document is first fed into an LED pre-trained model \protect\citep{beltagy2020longformer} (the upper half). Then, the acquired token-level attention graph (not shown here) is converted into a span-level  graph (the lower half) via the method described in Section \protect\ref{sec:span_linker}. Spans (which might be far apart) are then linked if their edge weight is high. For example, the span ``\textit{The main contributions}'' walks through 1,065 tokens and links with ``\textit{a single-layer forward recurrent neural network}'', which is then linked with ``Long Short-Term Memory'' since their high weight edges (0.53 and 0.48). Other spans do not connect with them due to their low edge weights to these spans.}
\end{figure}

Intuitively, the high cost of annotating long-document QA pairs can be alleviated in an unsupervised manner. However, there are only short-document unsupervised question answering (UQA) works \citep{DBLP:conf/acl/LewisDR19,DBLP:conf/naacl/PanCXKW21}, which aim to construct a large number of short-document QA pairs in an unsupervised manner and train a QA model with these QA pairs. \citet{DBLP:conf/acl/LewisDR19} first propose the UQA task and use unsupervised neural translation to construct QA pairs in a short passage. \citet{DBLP:conf/naacl/PanCXKW21} raise the unsupervised short-document multi-hop question answering (UMQA) task and design a question generation method to build multi-hop questions within two short passages. 
To break the document length limitation and incorporate long-range information, we propose a more challenging task, i.e. unsupervised long-document question answering (ULQA) task, to generate high-quality long-document QA pairs and train a competitive QA model without any human-labeled \textit{long}-document QA pairs. 

The core challenge of this task is in the modeling of long-range dependency without supervision. To address this issue, we study an \textit{attention-driven} method to incorporate meaningful long-range information in the constructed QA pairs. Figure \ref{fig:illustrator} illustrates a motivating example of the attention flow in a long document. It is observed that, by walking through the attention edges of a pre-trained model, related spans would be linked and long-range dependency in the document could be constructed. Therefore, long-range information could be also incorporated into QA pairs through these \textit{walkable} attention patterns among text spans. Thus, we propose \textit{AttenWalker}, a novel unsupervised framework to generate long-range dependent answers in long-document QA pairs. Specifically, AttenWalker comprises three modules: span collector, span linker and answer aggregator. Firstly, the span collector takes advantage of the constituent parsing and reconstruction ability of a pre-trained model to select informative candidate spans. Secondly, related spans that might be far apart could be connected through local or global attention edges of a long-document pre-trained model. Thirdly, collected spans are aggregated through the reconstruction ability of a pre-trained model.

Extensive experiments on Qasper \citep{DBLP:conf/naacl/DasigiLBCSG21} and NarrativeQA \citep{DBLP:journals/tacl/KociskySBDHMG18} show that the proposed AttenWalker can effectively model long-range dependency in long-document QA. Besides, AttenWalker also shows strong performance in the few-shot learning setting.

Our contributions are as follows:
\begin{itemize}
    \item To the best of our knowledge, we are the first to explore unsupervised \textit{long}-document QA.
    \item Without the human-annotated long-range knowledge, we propose AttenWalker, a novel unsupervised long-document QA framework, which can incorporate long-range reasoning via attention-based graph walking.
    \item Extensive experiments show that AttenWalker outperforms previous methods in unsupervised and few-shot settings.
\end{itemize}

\section{Related Works}
\paragraph{Unsupervised Question Answering}
Unsupervised question answering (UQA) \citep{DBLP:conf/acl/LewisDR19} targets at alleviating the data scarcity problem in QA datasets. It focuses on generating QA pairs without supervision and training a QA model on them. \citet{DBLP:conf/acl/LewisDR19} firstly propose the UQA task. Based on a pure short document, they extract answers via named entity tools and propose a novel cloze translation method to make alignment between cloze question and natural question so as to generate plenty of natural questions. Then, the constructed (context, question, answer) triples are used to train a QA model. \citet{DBLP:conf/acl/LiWDWX20} use cited documents to generate questions so that the overlapping problem between the generated question and the raw context could be alleviated. \citet{DBLP:conf/coling/NieHCM22} propose to mine answers beyond named entities in the synthetic QA dataset and improve the model's ability in dealing with diverse answers. \citet{DBLP:conf/naacl/PanCXKW21} propose the first unsupervised multi-hop QA framework via multi-hop question generation. However, most of these methods focus on the short-document scenario, while the long-document setting is still unexplored.

\begin{figure*}[t]
\centering
\centerline{\includegraphics[width=.8\textwidth]{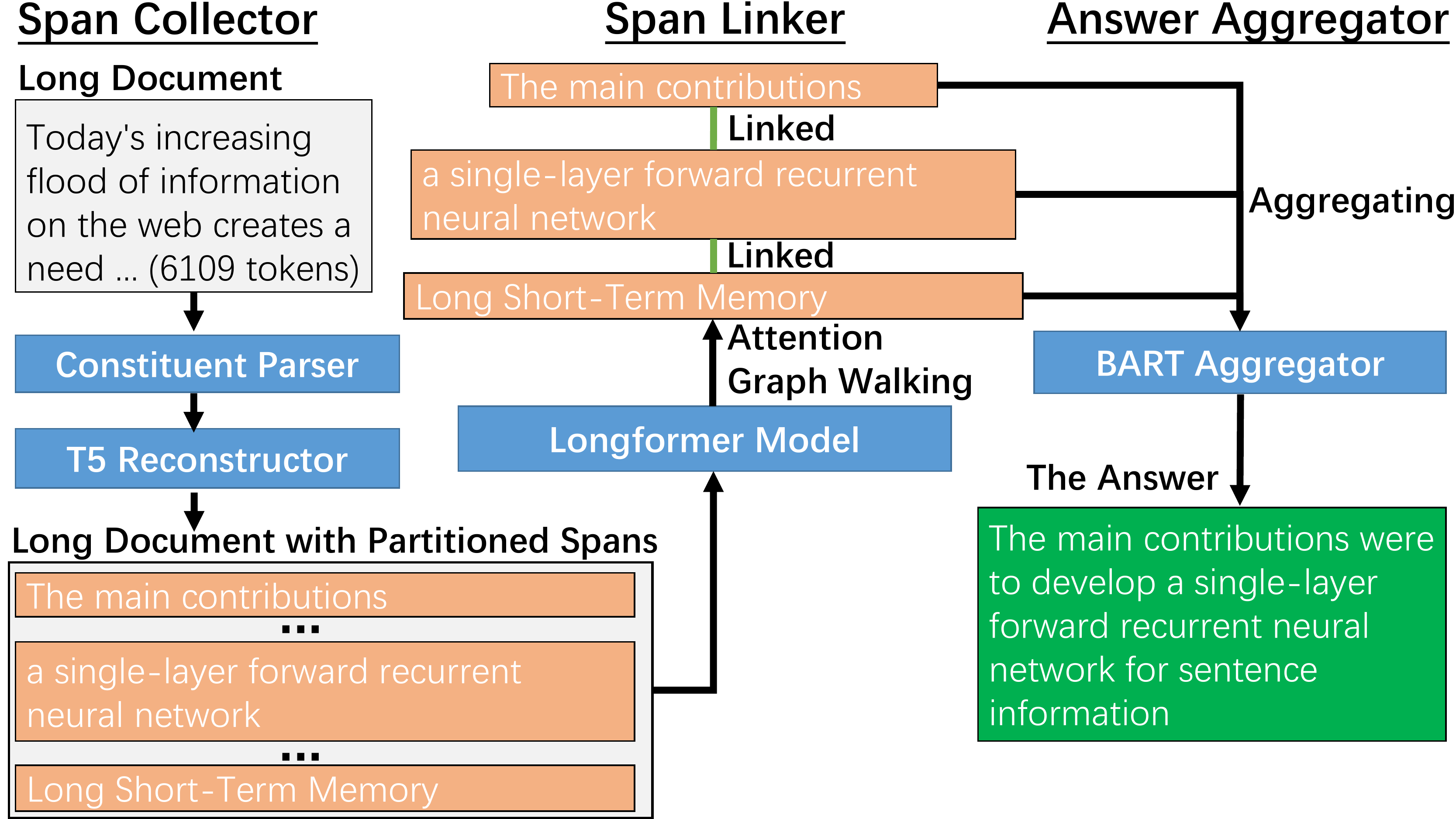}}
\caption{\label{fig:model}An overview of AttenWalker. It consists of three modules, including Span Collector, Span Linker, and Answer Aggregator.}
\end{figure*}
\paragraph{Long-document Question Answering} Long-document question answering (long-document QA) aims to answer questions based on the understanding of a long sequence of text. Previous methods can be divided into end-to-end methods and select-then-read methods. End-to-end methods \citep{DBLP:conf/naacl/DasigiLBCSG21} apply sparse attention models to directly answer the question given a long document. \citet{DBLP:conf/naacl/DasigiLBCSG21} uses the Longformer-Encoder-Decoder model to make long-range reasoning on a long document and then answer a question. \citet{caciularu-etal-2022-long} uses a sequence-level objective to improve evidence verification. For the select-then-read methods, \citet{nie2022capturing} propose a compressive graph selector network to select question-related snippets from the long document and then use the selected short snippets for answer generation. However, despite competitive performances on long-document QA, these methods heavily rely on supervised QA data and can hardly apply to the low-resource setting.

\section{AttenWalker}
In this section, we first formalize the task of long-document QA. After that, the proposed AttenWalker is described in detail. 

\subsection{Problem Formulation}
The setup of long-document QA is as follows. Given a question $q$ and a long document $c$, where $c$ is often more than 10K tokens, the QA model $p_{\theta}(a|c,q)$ needs to produce a free-formed answer $a$ by understanding the long document $c$ and aggregating question-related snippets from $c$.

In this paper, we consider an unsupervised setting, where only long document $c$ is available. Our aim is to generate synthetic QA pairs $(q', a')$ with long-range information and train a competitive long-document QA model via ($c,q',a'$) triples.

\subsection{Overview of the Method}

The proposed AttenWalker focuses on incorporating long-range information via a well-designed answer generator. Specifically, AttenWalker comprises three modules: Span collector, Span linker, and Answer Aggregator. As shown in Figure \ref{fig:model}, the Span Collector first partitions the Long Document into different spans via Constituent Parsing and T5 Reconstructor. Secondly, a Span Linker is used to capture long-range dependency among these Partitioned Spans via Attention Graph Walking. This module aims to walk through local and global attention edges to link semantically related spans (which could be far apart in the long text) for aggregating answers. Thirdly, an Answer Aggregator combines all the Linked Spans via the reconstruction ability of a BART model to generate the answer.

\begin{figure}
\centering
\includegraphics[width=\linewidth]{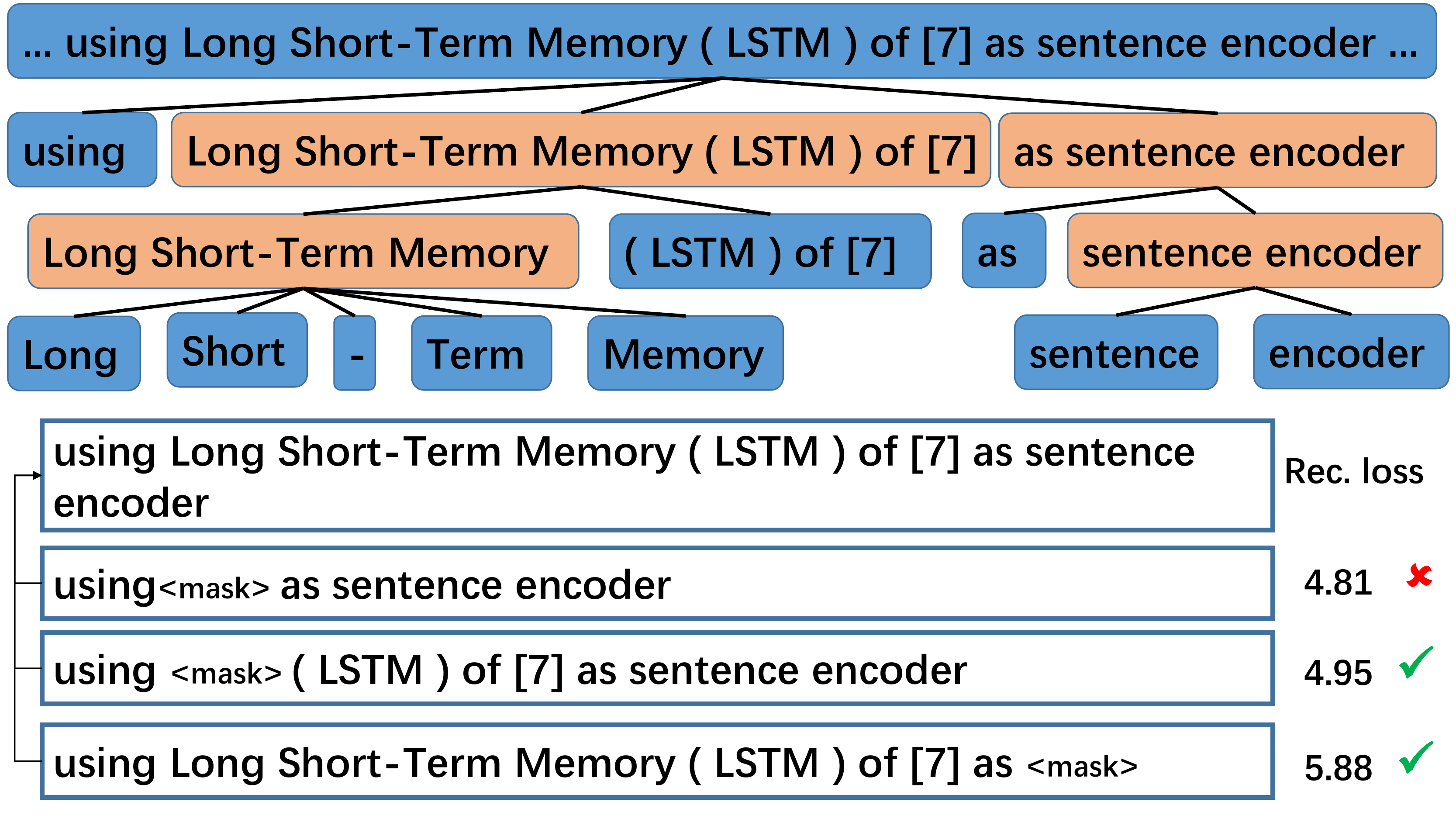}
\vspace{-0.3cm}
\caption{\label{fig:answer_collector}An illustration of the Span Collector of AttenWalker. To determine informative spans in the sentence ``\textit{using Long Short-Term Memory (LSTM) of [7] as sentence encoder}'', the Constituent Parser first partitions the sentence into spans.
These spans are then masked out and a pre-trained model (T5 Reconstructor) is used to make a reconstruction, where higher reconstruction loss could indicate a more informative span.}
\vspace{-0.3cm}
\end{figure}

\subsection{Span Collector}
\label{sec:span_collector}
To determine the candidate spans for generating the answers, we propose a \textit{Span Collector}. Specifically, as shown in Figure \ref{fig:answer_collector}, it first seeks for candidate spans via constituent parsing and then reconstructs masked text via a pre-trained T5 model \citep{DBLP:journals/jmlr/RaffelSRLNMZLL20} to select informative spans for answer generation. 
Each masked text serves as an input to the T5 model\footnote{In practice, we use \texttt{<extra\_id\_0>} as the mask token. The \texttt{<mask>} token in Figure \ref{fig:answer_collector} is just for illustrative purpose.}. The reconstruction loss is:
\begin{equation}
    \mathcal{L}=-\frac{1}{T}\sum_{i=1}^T\text{log}(p(y_i))~,
\end{equation}
where $\mathcal{L}$ is the reconstruction loss of the specific span. $T$ is the number of tokens in the ground truth span and $p(y_i)$ is the T5 predicting probability of the $i$-th token $y_i$ in the ground truth span. As shown in Figure \ref{fig:answer_collector}, ``\textit{sentence encoder}'' has the largest reconstruction loss. Thus, we select it as one of the candidate spans. Meanwhile, its parent spans (i.e. ``\textit{as sentence encoder}'') and its child spans (``\textit{sentence}'' and ``\textit{encoder}'') will not be selected for redundancy concern. 

\subsection{Span Linker}
\label{sec:span_linker}

\begin{figure}[t]
\centering
\includegraphics[width=\linewidth]{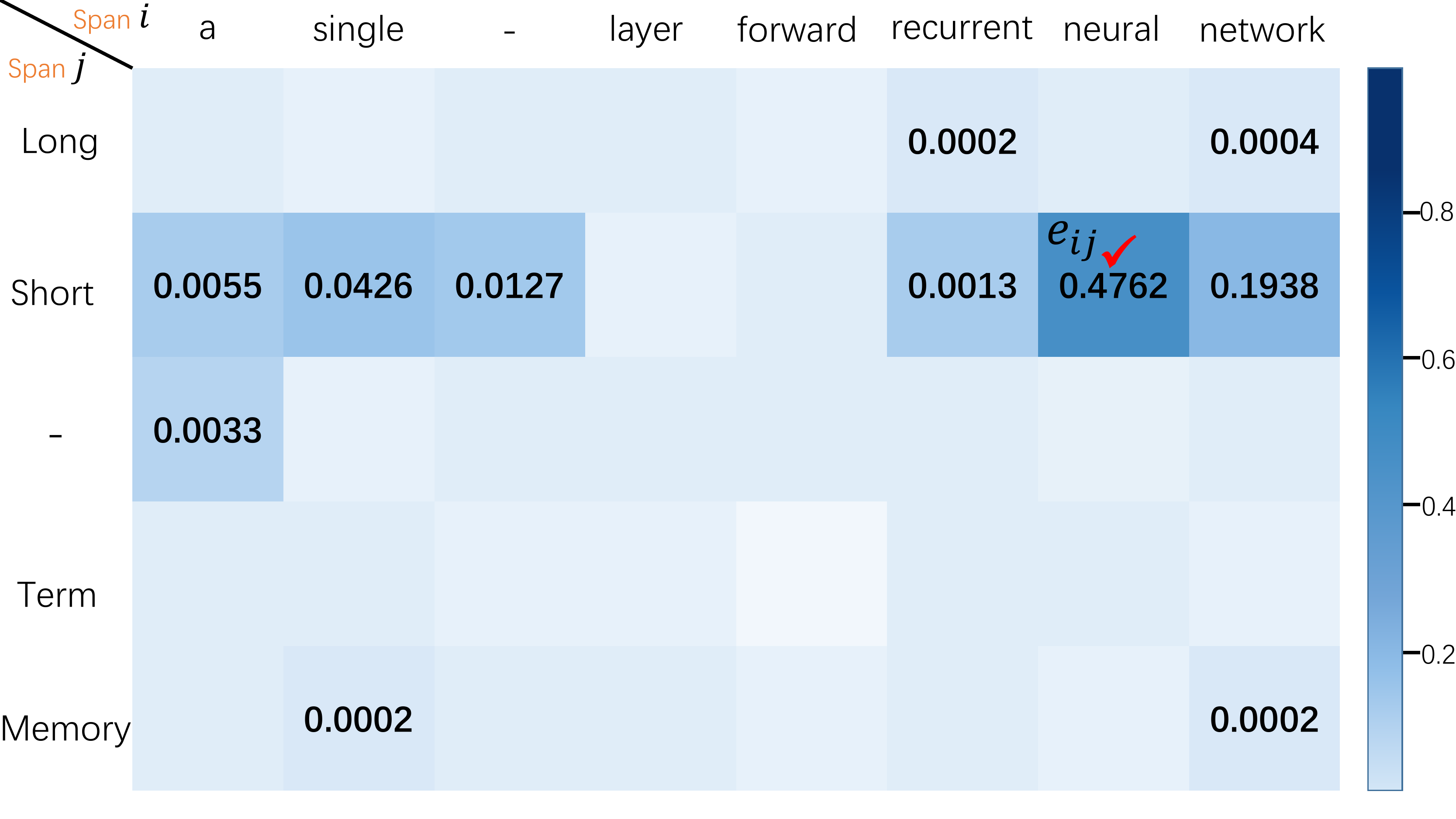}
\caption{\label{fig:max_pooling} An attention heatmap from each token in span ``\textit{a single-layer forward recurrent neural network}'' to each token in span ``\textit{Long Short-Term Memory}''. Attention score values lower than 0.0001 are not displayed. The highest value 0.4762 is selected as the edge weight between these two spans.}
\end{figure}

The proposed Span Linker is to incorporate long-range information in AttenWalker. It can effectively incorporate long-range dependency through attention-based graph walking. The Span Linker is composed of two sub-modules: a Span Graph Constructor and an Attention-based Graph Walker.

\paragraph{Span Graph Constructor}
To explore possible relations among spans, token-level
attention scores\footnote{The token-level attention scores are acquired through the encoder part of the LED model. We consider each span graph for each Transformer layer and head.} of the LED pre-trained model \citep{beltagy2020longformer} can be used. As shown in Figure \ref{fig:illustrator}, based on the spans acquired in Section \ref{sec:span_collector}, we build a span graph $\mathcal{G}$ via attention scores between each pair of tokens as shown in Figure \ref{fig:max_pooling}. For span $i$ and span $j$, where $i,j\in\mathcal{G}$, if there are any attention edges from one of the tokens in span $i$ to one of the tokens in span $j$, there is an edge from span $i$ to span $j$. Motivated by the idea of max-pooling technique \citep{dumoulin2016guide}, to obtain the most obvious relation in each pair of spans, the edge weight $e_{ij}$ from span $i$ to span $j$ can be calculated by the maximum attention weight between any pair of tokens in between:
\begin{equation}
e_{ij}=\underset{m\in\mathcal{G}_i,n\in\mathcal{G}_j, (m,n)\in\mathcal{G}_t}{\max}\ w_{m,n}~,
\label{eqn:max_pooling}
\end{equation}
where $\mathcal{G}_i$ and $\mathcal{G}_j$ are tokens in span $i$ and span $j$. $(m,n)$ is an edge in token-level attention graph $\mathcal{G}_t$. $w_{m,n}$ is the attention weight of the edge $(m,n)$. 

In the LED encoder, there are local and global attention weights among the tokens in a long document. Both two types of weights can serve as the token-level edge weights $w_{m,n}$ in Eqn \ref{eqn:max_pooling}. In this work, we propose to consider both types for span graph construction. If there is a local attention weight $l_{m,n}$ from token $m$ to token $n$, we directly assign the value to $w_{m,n}$. Otherwise, the global attention is considered: we insert a ``\texttt{</s>}'' at the beginning of each paragraph and set global attention for each of it (Appendix \ref{appendix:details_in_fine_tuning_the_led_model}). It means that each ``\texttt{</s>}'' can attend to every token in the long sequence and vice versa. Each ``\texttt{</s>}'' could serve as the representative of the paragraph that follows it. Therefore, ``\texttt{</s>}'' can be regarded as a bridge to two spans in different paragraphs, which could be far apart and could not be accessible to each other only through the local attention mechanism. To build the ``bridge'' from paragraph $p_i$ to paragraph $p_j$, we first select one of the K tokens ${t_{p_i}}$ with the maximum attention score to the representation of ``\texttt{</s>}'' $s_{p_i}$. Next, for the representation of $s_{p_i}$, $L$ highest attention scores to other ``\texttt{</s>}'' tokens are selected. For one of the L ``\texttt{</s>}'' tokens $s_{p_j}$ in paragraph $p_j$ , we can access its maximum $M$ attention weights to the corresponding $M$ tokens (${t_{p_j}}$) in paragraph $p_j$. For each  ${t_{p_i}}$, its attention to the target token ${t_{p_j}}$ can be:
\begin{equation}
    g_{{t_{p_i}},{t_{p_j}}} = \sqrt[3]{w_{{t_{p_i}},{s_{p_i}}}\times w_{s_{p_i},s_{p_j}}\times ,w_{s_{p_j},{{t_{p_j}}}}}~,
\end{equation}
where $g_{t_{p_i},t_{p_j}}$ is the global attention score from token $t_{p_i}$ to token $t_{p_j}$. $w_{{t_{p_i}},{s_{p_i}}},w_{s_{p_i},s_{p_j}},w_{s_{p_j},{{t_{p_j}}}}$ are attention scores directly acquired according to the global attention in the LED model. Here, we use the geometric mean of the attention edge weights from $t_{p_i}$ to $t_{p_j}$ as the approximate attention weight of the edge $(t_{p_i}, t_{p_j})$. Thus, if there is no direct (local) attention from $t_{p_i}$ to $t_{p_j}$ but a global path, we can use $g_{t_{p_i},t_{p_j}}$ as the ``lost'' $w_{t_{p_i},t_{p_j}}$. 

\begin{figure}[t]
\centering
\includegraphics[width=\linewidth]{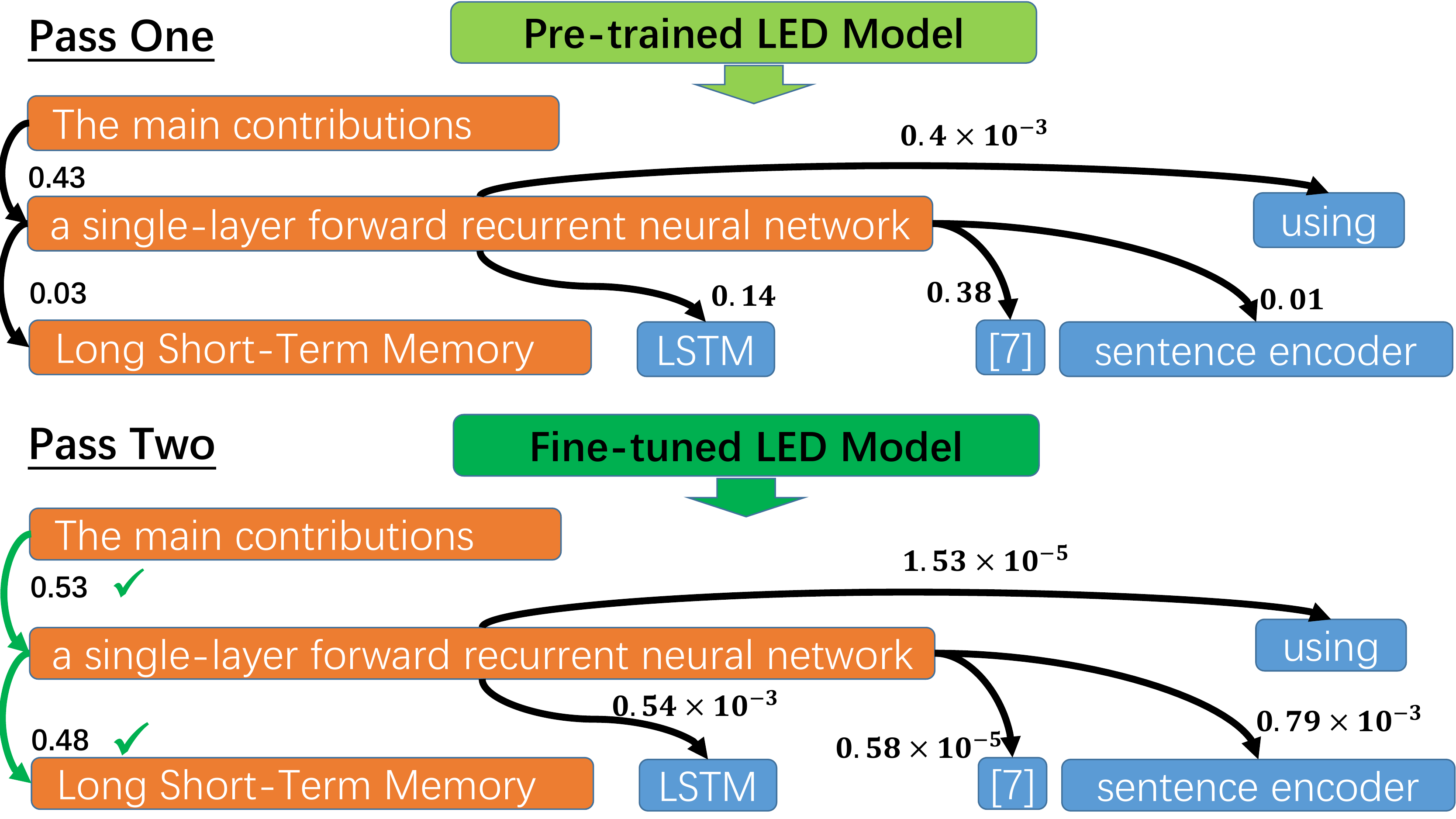}
\caption{\label{fig:answer_linker} An example document (in Qasper training set) of edge weight changes from the first pass to the second pass.}
\end{figure}

\paragraph{Attention-based Graph Walker}
Span linking can be done via attention-based graph walking on the constructed span graph. Essentially, the proposed graph walker collects interrelated spans via traversing the span graph. Its main algorithm is based on the Depth First Search \citep{even2011graph}. As shown in the lower half of Figure \ref{fig:answer_linker}, starting from the first span ``\textit{The main contributions}'', graph walking continues searching for accessible span. Thus, it successfully links to the span ``a single-layer forward recurrent neural network''. Then, starting from this linked span, ``Long Short-Term Memory'' is also linked because of the high weight 0.48 between it and ``\textit{a single-layer forward recurrent neural network}''. To decide whether the edge is of ``high weight'', we set a pre-defined threshold $\tau$ on the edge weight.  In other words, the original span graph $\mathcal{G}$ can be pruned as a new graph $\mathcal{G}'$ via:
\begin{equation}
\mathcal{G}'=\left\{e|e\in\mathcal{G},w_e>\tau\right\}~,
\end{equation}
where $w_e$ is the weight of edge $e$.
Finally, spans on the walking path are clustered together, which will be used in the following section.

\subsection{Answer Aggregator}
\label{sec:answer_aggregator}
The proposed Answer Aggregator produces the final answer by aggregating the linked spans in Section \ref{sec:span_linker}.
To achieve this goal, we take advantage of the reconstruction ability of a BART model \citep{lewis2020bart}. For instance, the linked spans in the lower half of Figure \ref{fig:answer_linker} can be formalized into the input to BART: ``\texttt{The main contributions <mask> a single-layer forward recurrent neural network <mask> sentence information}''. Finally, the output can be an integral text as the answer: ``\texttt{The main contributions were to develop a single-layer forward recurrent neural network for sentence information}''.
\begin{figure}[t]
\centering
\includegraphics[width=\linewidth]{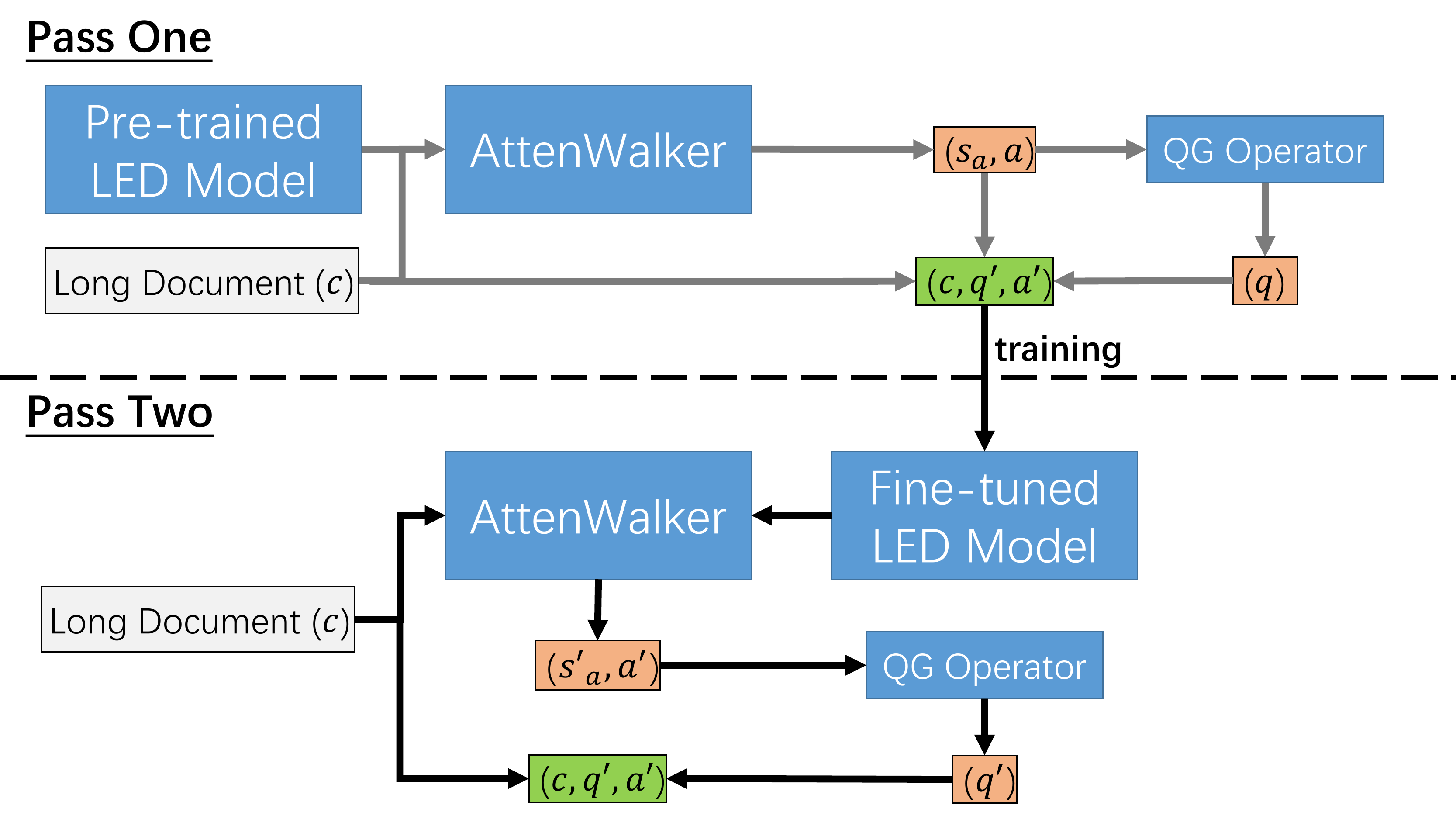}
\caption{\label{fig:two_pass_scheme} Overview of the proposed two-pass scheme. $s_a,{s_a}'$ are the sentences of the linked spans for $a,a'$}
\end{figure}

\begin{table*}[t]
\fontsize{10.0}{10.0}\selectfont
    \centering
    \begin{tabular}{lccc|cccc}    
        \toprule
        \multirow{2}{*}{Models} & \multicolumn{3}{c}{Qasper} & \multicolumn{4}{c}{NarrativeQA}\\
        & Extractive & Abstractive & Overall & Bleu-1 & Bleu-4 & Meteor & Rouge-L\\
        \midrule
        \multicolumn{4}{l}{\textit{Supervised}}\\
        LED \citep{DBLP:conf/naacl/DasigiLBCSG21}  & 30.92 & 14.91 & 26.05& 20.04 & 2.34 & 6.43 & 16.16\\
        \ \ \ \ \ + MQA-QG  & 28.98 & 13.87 & 24.42& 20.88 & \textbf{3.35} & 6.99 & 17.38\\
        \ \ \ \ \ + AttenWalker & \textbf{32.44} & \textbf{15.41} & \textbf{27.08}& \textbf{21.15} & 2.99 & \textbf{7.03} & \textbf{18.07}\\
        Human & 58.92 & 39.71 & 52.80& 44.43 & 19.65 & 24.14 & 57.02\\ 
        \midrule
        \multicolumn{4}{l}{\textit{Unsupervised}}\\
        UNMT \citep{DBLP:conf/acl/LewisDR19}& 6.72 & 2.78 & 4.13& 5.68 & 0.00 & 1.03 & 3.82 \\
        RefQA \citep{DBLP:conf/acl/LiWDWX20}& 3.08 & 0.63 & 2.26& 0.95 & 0.00 & 1.02 & 0.96 \\
        DiverseQA \citep{DBLP:conf/coling/NieHCM22}& 5.35 & 4.69 & 5.13& 0.79 & 0.00  & 1.14 & 1.03\\
        MQA-QG \citep{DBLP:conf/naacl/PanCXKW21}& 11.88 & 5.91 & 9.85& 6.65 & 0.00 & 1.90 & 4.38\\
        AttenWalker & \textbf{17.21} & \textbf{12.66} & \textbf{15.72} & \textbf{9.39} & \textbf{0.91} & \textbf{3.82} & \textbf{7.71}\\
        \bottomrule   
    \end{tabular}
    \caption{\label{tab:main_results} The performance on the test set of Qasper and NarrativeQA. In the second row, ``Extractive, Abstractive, Overall'' refer to Extractive F1, Abstractive F1 and Overall F1 in Qasper. In the ``\textit{Supervised}'' block, the row ``LED'' denotes the performance of an LED model fine-tuned on the supervised dataset. ``+MQA-QG'' means that an LED model is first trained on the synthetic QA pairs from MQA-QG, and then continuously trained on supervised data. The meaning of ``+AttenWalker'' is similar. In the ``\textit{Unsupervised}'' block, each unsupervised method generates long-document QA pairs and an LED model is fine-tuned on them without any supervised QA instances.}
\end{table*}

\subsection{Question Generation}
Question generation (QG) is applied when we obtain the answer and all the sentences the linked spans from. We use the QG Operator in Unsupervised Multi-hop QA \citep{DBLP:conf/naacl/PanCXKW21} as the QG module in our work. We concatenate the answer from Section \ref{sec:answer_aggregator} with all the aforementioned sentences into the QG module to generate a question.

\subsection{Two-Pass Scheme for Long-Range Reasoning}
In the pre-trained LED model, query, key, and value matrices of the global attention are just copied from the corresponding matrices in the local attention\footnote{\url{https://github.com/allenai/longformer}}. To further improve the ability of global attention in long-range reasoning, we design a two-pass scheme to construct long-document QA pairs as shown in Figure \ref{fig:two_pass_scheme}. In the first pass, only local attention is used in the proposed Span Graph Constructor. Then,  an LED model is fine-tuned on these QA pairs with global and local attention as described in Appendix \ref{appendix:details_in_fine_tuning_the_led_model}. This step aims to improve the ability of the query, key, and value matrices, especially for global attention. In the second pass, based on the fine-tuned LED model, both local and global attention are considered to construct the span graph for attention walking. Hence, further knowledge with global attention is incorporated into the finally constructed QA pairs.

\section{Experimental Setup}
We evaluate the proposed AttenWalker on Qasper \citep{DBLP:conf/naacl/DasigiLBCSG21} and NarrativeQA \citep{DBLP:journals/tacl/KociskySBDHMG18}. In particular, for Qasper, the answer types in this dataset can be extractive, abstraction, yes/no, or unanswerable. Yet, according to our analysis (Appendix \ref{appendix:analysis_of_qasper_question_types}), QA instances with yes/no or unanswerable answers cannot properly evaluate the ability of long document reasoning. Therefore, we only focus on the extractive and abstractive QA instances in this work. The datasets splitting and processing details are in appendix \ref{appendix:sec:datasets}.

We use the documents in the Qasper training set to construct QA pairs for training the QA model and do Qasper-related experiments. The long documents in the training set of NarrativeQA are used similarly. The dataset construction details can be found in Appendix \ref{appendix:sec:unsupervised_long_document_qa_dataset_construction}. What's more, the setting of the long document QA model trained on the constructed dataset can be referred to \ref{appendix:sec:long_document_qa_model_setting}.

\begin{table*}[t] 
\fontsize{10.5}{10.8}\selectfont
    \centering
    \begin{tabular}{lccc|cccc}    
        \hline
        \multirow{2}{*}{Datasets} & \multicolumn{3}{c|}{Qasper}& \multicolumn{4}{c}{NarrativeQA}\\
        & Extractive & Abstractive & Overall & Bleu-1 & Bleu-4 & Meteor & Rouge-L\\
        \hline
        AttenWalker & 12.13& \textbf{15.57}& \textbf{13.28}& \textbf{9.62}& \textbf{1.11}& \textbf{3.83}& \textbf{7.39}\\
        w/ Random Span Collector & 9.06& 8.65& 8.93& 8.40& 0.67& 2.67& 6.25\\
        w/ Un-pre-trained LED & 9.39& 7.80& 8.90& 0.59& 0.00& 1.11& 0.93\\
        w/ Embedding Linker & 11.69& 9.04& 10.87& 6.33& 0.24& 2.87& 5.60\\
        w/o Global & 11.36& 10.75& 11.16& 7.23& 0.38& 3.10& 6.06\\
        w/ Answer Connector & 9.48& 10.00& 9.66 & 6.66& 0.00& 3.13 & 5.99\\
        w/ Single Pass & \textbf{12.52}& 10.99& 12.02& 7.77& 0.62& 3.33& 6.60\\
        w/ Single Pass + Global& 12.07& 11.25& 11.81& 7.55& 0.34& 2.94& 5.66\\
        \hline   
    \end{tabular}
    \caption{\label{tab:ablation} Ablation study of AttenWalker, evaluating on the dev set of Qasper and NarrativeQA. ``w/ Random Span Collector'' denotes that candidate spans are randomly selected. ``w/ Un-pre-trained LED'' uses an LED model with randomly initialized parameters in the Span Linker. ``w/ Embedding Linker'' calculates attention scores only by the inner-product values between each pair of input embeddings. ``w/o Global'' does not consider the global attention in AttenWalker. ``w/ Answer Connector'' directly connects linked spans to form the answer. ``w/ Singe Pass'' only uses the pass-one in the proposed Two-Pass Scheme, while ``w/ Single Pass + Global'' further add global attention in it.} 
\end{table*}

\section{Experiment}
In this section, we first discuss the main results of AttenWalker on Qasper and NarrativeQA, and then further analyze the proposed method.

\subsection{Main Results}
Since there is no direct unsupervised method for long documents, we select competitive baselines from unsupervised short-document QA (UQA) and unsupervised short-document multi-hop QA (UMQA). The UQA works include UNMT \citep{DBLP:conf/acl/LewisDR19}, RefQA \citep{DBLP:conf/acl/LiWDWX20}, DiverseQA \citep{DBLP:conf/coling/NieHCM22}. The UMQA work is MQA-QG \citep{DBLP:conf/naacl/PanCXKW21}. The adaptation of them to long documents is described in Appendix \ref{appendix:details_in_implementing_baselines}. Following \citet{DBLP:conf/naacl/DasigiLBCSG21} and \citet{DBLP:journals/tacl/KociskySBDHMG18}, we use answer F1 score (including extractive F1, abstractive F1 and overall F1 in this paper) as the evaluation metrics
on Qasper dataset, while we use Bleu-1/4 \citep{papineni2002bleu}, Meteor \citep{denkowski2011meteor} and Rouge-L \citep{lin2004rouge} for evaluation on NarrativeQA dataset.

As shown in Table \ref{tab:main_results}, in the \textit{Supervised} block, it can be found that an LED model trained on the synthetic dataset of AttenWalker can further make improvements when it is continuously fine-tuned on the supervised data, especially on Qasper, showing that the proposed method can effectively alleviate the data scarcity problem in Qasper. In the \textit{Unsupervised} block, the proposed AttenWalker outperforms all baselines by a large margin in the fully unsupervised setting. showing a competitive performance of AttenWalker.

\subsection{Ablation Study}
We conduct an extensive ablation study on different components of AttenWalker. As shown in Table \ref{tab:ablation}, the effectiveness of each component can be shown according to four observations.

\paragraph{Effects of the span collector.}
As shown in Table \ref{tab:ablation}, the performance drop of ``w/ Random Span Collector'' illustrates that randomly selecting candidate spans could introduce much noise and harm the quality of the generated QA pairs. 
\begin{figure}[t]
\centering
\includegraphics[clip,width=.4\textwidth]{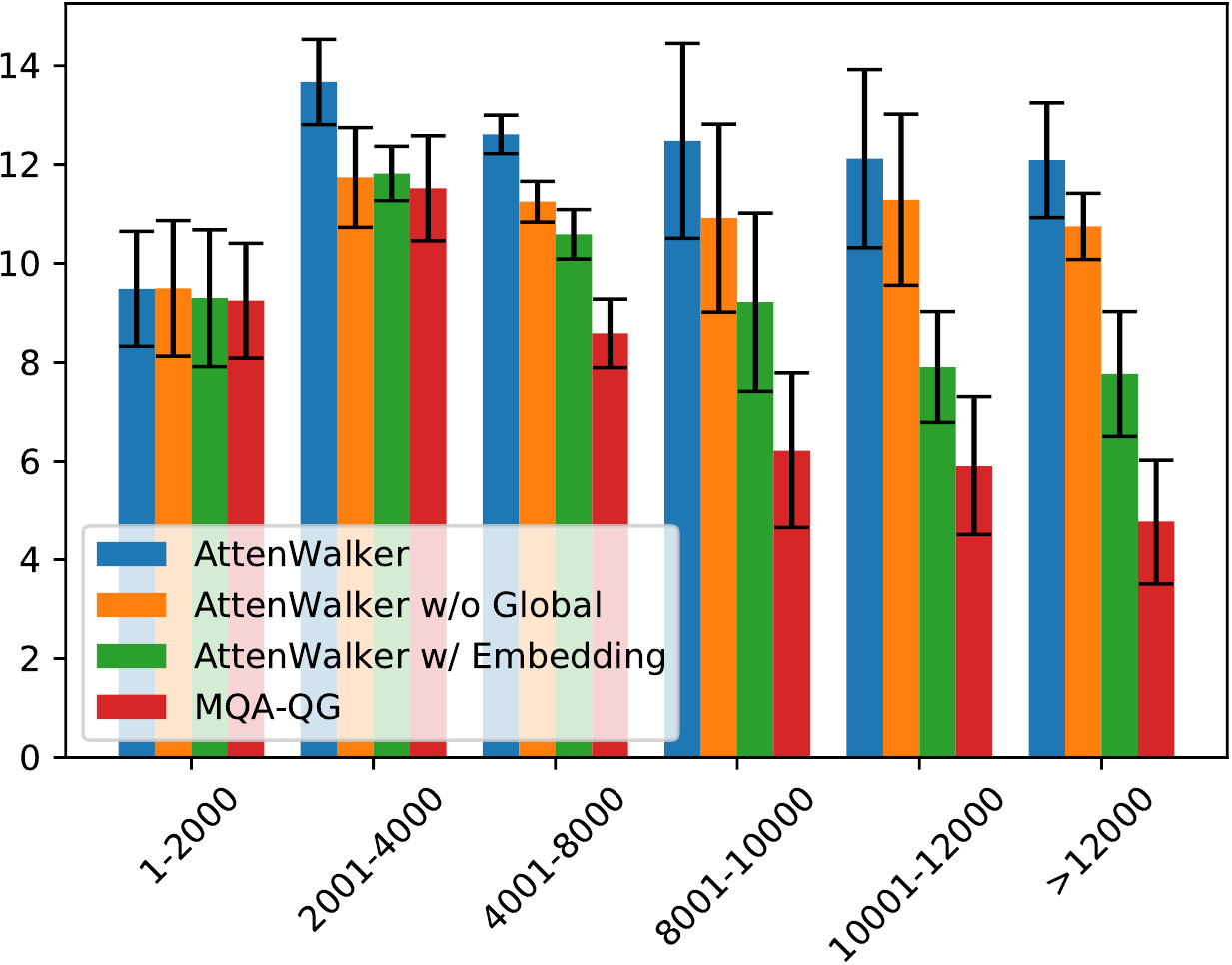}%
\caption{\label{fig:doc_len}The mean and variance of Overall F1 (with 5 random seeds) for AttenWalker, two ablated versions ``w/o Global''``w/ Embedding'' and MQA-QG. The dev set of Qasper is divided based on document length.}
\end{figure}

\paragraph{Effects of the span linker.}
From the performance drop in setting ``w/ Un-pre-trained LED'' and ``w/ Embedding Linker'' as shown in Table \ref{tab:ablation}, it can be known that the attention information stored in the LED parameters is rather useful for constructing high-quality long-document QA pairs. Besides, the competitive result of ``w/ Embedding Linker'' suggests that embedding information can benefit the QA pair construction. In addition, the performance of ``w/o Global'' illustrates that global attention is also an essential factor in improving the quality of the generated long-document QA pairs.

\paragraph{Effects of the answer aggregator.}
According to ``w/ Answer Connector'' in Table \ref{tab:ablation}, the performance drops when simply connecting spans. It shows that connecting spans with proper transition words is crucial for generating a high-quality answer.

\paragraph{Effects of the two-pass scheme.}
The Two-Pass Scheme is helpful in improving the performance of the model as shown in the ``w/ Single Pass'' and ``w/ Single Pass + Global'' setting from Table \ref{tab:ablation}. It suggests that local and global attention can benefit from the parameters of a fine-tuned LED model.
\subsection{Effects on Long-Range Modeling}
\label{sec:effects_on_long_range_modeling}
AttenWalker aims to incorporate long-range information in the QA pair construction. To further understand it, an experiment with varied document lengths is conducted. As shown in Figure \ref{fig:doc_len}, in essence, ``w/o Global'' is only to use local attention while ``w/ Embedding'' denotes a situation that both global and local are not used. When the document length is small (1-2,000), the performances of different methods are comparable. However, with the increasing document length, the gap among methods becomes larger. It shows that AttenWalker can model long-range dependency effectively. Furthermore, it is observed that MQA-QG performs worse than ``w/ Embedding'' when the document length is large. It can be explained in two aspects. Firstly,  MQA-QG could hardly capture long-range information. Secondly, MQA-QG is only a reduced version of ``w/ Embedding'', which can only link two spans via literal matching (Section \ref{sec:case_study}).
\subsection{Effects of Attention Weights}

We design three different span graph construction strategies to further investigate their influences on the proposed method. As shown in Table \ref{tab:pooling}, the ``Max-Pooling'' strategy outperforms the other two strategies by large margins. It can be explained that the ``Max-Pooling'' strategy can capture the most obvious (and probably important) relation between two spans, which is useful in QA pair construction.

\begin{table}[t] 
\small
    \centering
    \begin{tabular}{lccc}    
        \hline
         & Extractive & Abstractive & Overall \\
        \hline
        Max-Pooling* & \textbf{12.13}& \textbf{15.57}& \textbf{13.28}\\
        Min-Pooling & 6.79& 5.78& 6.47\\
        Mean-Pooling & 6.81& 6.42& 6.54\\
        \hline   
    \end{tabular}
    \caption{\label{tab:pooling} Overall F1 of several methods with different strategies to build span graph, on the Qasper dev set. ``Max-Pooling*'' is used in AttenWalker, where the maximum attention score between tokens of two spans is selected as the edge weight. Similarly, ``Min-Pooling'' uses the minimum attention score, while ``Mean-Pooling'' uses the average of attention scores.} 
\end{table}

\begin{figure*}[t]
\centering

\includegraphics[clip,width=.9\columnwidth]{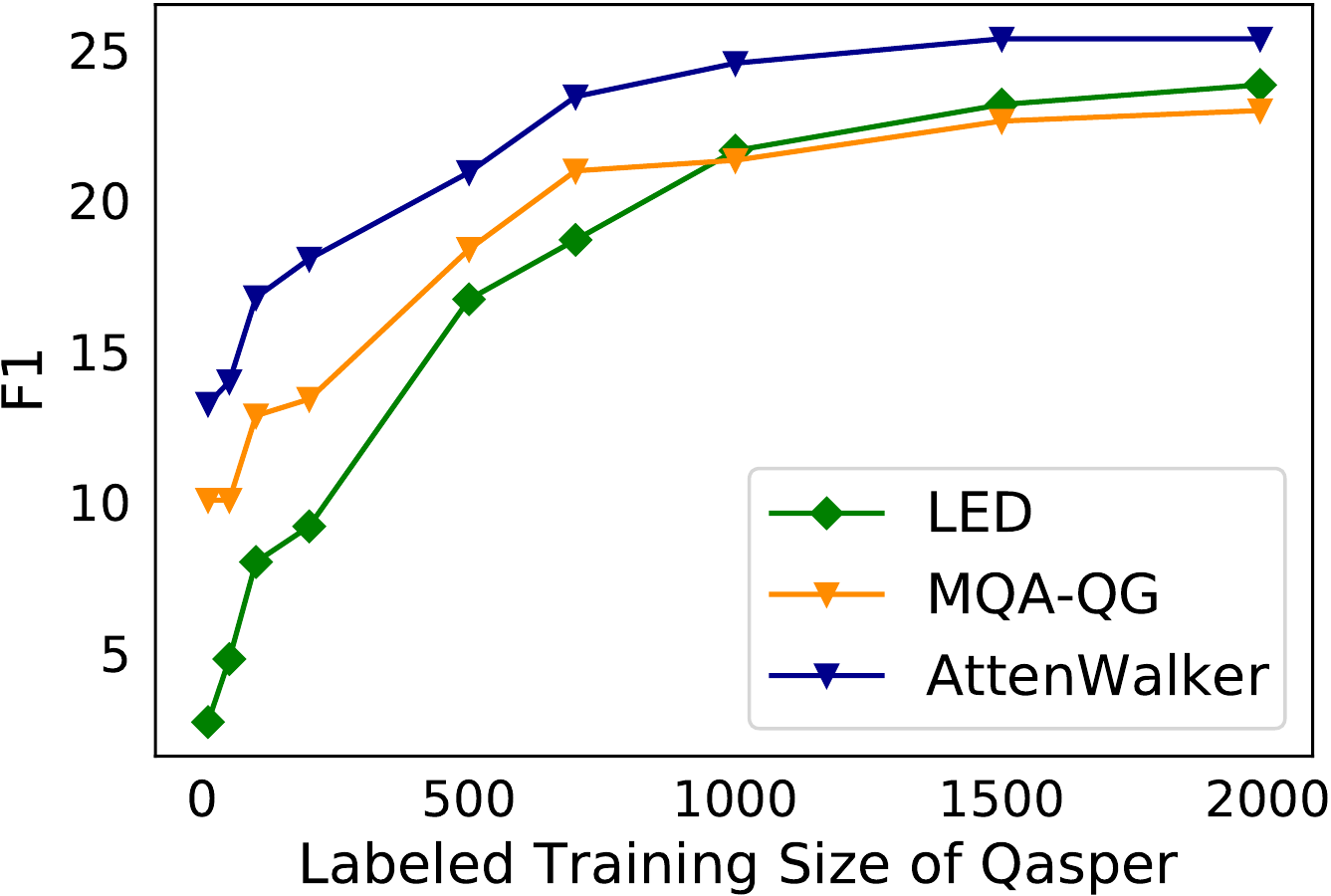}%
\includegraphics[clip,width=.9\columnwidth]{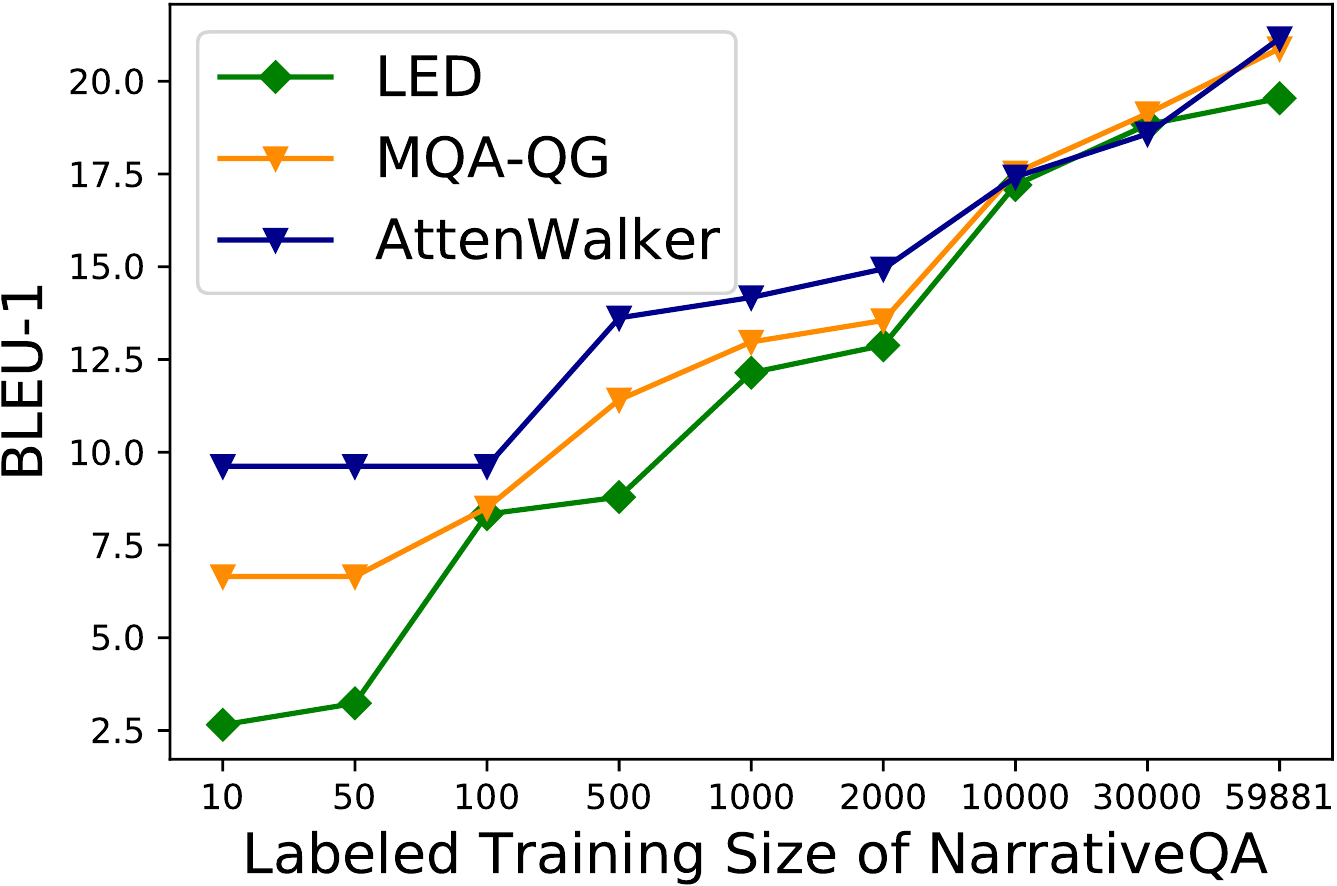}%

\caption{\label{fig:few_shot}The few-shot learning of three methods on different sizes of labeled training data, evaluated on the dev set.}

\end{figure*}
\begin{table*}[t] 
\small
    \centering
    \begin{tabular}{p{.95\textwidth}}   
        \toprule
        AttenWalker\\
        \midrule
         \textbf{Related Context:}
         ...... \textcolor{blue}{QG research} traditionally considers ...(1,909 tokens)... most commonly considered factor by current NQG systems is the \textcolor{blue}{target answer} ...(1,919 tokens)... the answer also deserves  \textcolor{blue}{more attention} from the  model...\\
         \textbf{Generated Answer}: QG research  shows the target answer  deserves  more attention\\
         \textbf{Generated Question}: What is the most commonly considered factor by current NQG systems?\\
        \midrule
        MQA-QG\\
        \midrule
         \textbf{Related Context:}
         ... They both follow the traditional decomposition of QG into \textcolor{blue}{content selection} and question construction ...(8 tokens)... For \textcolor{blue}{content selection}, [58] learn a sentence selection task to identify question-worthy sentences ...\\
         \textbf{Generated Answer}: content selection\\
         \textbf{Generated Question}: What is the task of identifying question-worthy parts in traditional the question that is the purpose of Question Generation synonymous with?\\
        \bottomrule   
    \end{tabular}
\caption{\label{tab:case_study}Examples of the generated QA instances from AttenWalker and MQA-QG given the same long document. Blue texts are selected spans for answer generation.} 
\end{table*}

\subsection{Few-Shot Learning}
We conduct the few-shot learning experiment to explore the effectiveness of AttenWalker in different low-resource settings. As shown in Figure \ref{fig:few_shot}, with the increasing of the labeled training size, the performance of the model trained on the synthetic QA pairs from AttenWalker is consistently better than that of MQA-QG in Qasper and an LED model. It is because the Qasper dataset is quite small, which makes the synthetic dataset rather beneficial. Besides, in the NarrativeQA, AttenWalker reaches the best performance from 10 to 10,000 training sizes and then becomes comparable with MQA-QG. It can be explained that a large number of training sizes would narrow the gaps between them. 

\subsection{Case Study}
\label{sec:case_study}
In this section, we first analyze an example with the proposed two-pass scheme to explore the benefits of attention changes. Then, we compare two QA examples between AttenWalker and MQA-QG. 

As shown in Figure \ref{fig:answer_linker}, with an LED model, the spans ``The main contributions'' can be connected with ``a single-layer forward recurrent neural network'' and ``[7]''. Yet, after fine-tuning the model with generated QA instances, a more reasonable path ``The main contributions'' -> ``a single-layer forward recurrent neural network'' -> ``Long Short-Term Memory'' is strengthened and the link to the trivial span ``[7]'' is weakened. It can be explained that after fine-tuning, noise in the LED attention edges is reduced, further improving the span linking and the quality of the generated QA instances.

In addition, as shown in Table \ref{tab:case_study}, we compare two QA pairs generated by AttenWalker and the best-performed baseline, MQA-QG. There are three key observations from the table. Firstly, AttenWalker can synthesize multiple spans into an answer whereas MQA-QG can only link the repeated text. Secondly, MQA-QG fails in long-range modeling since repeated spans could probably be in a short distance. Thirdly, the generated answer by AttenWalker is much more informative than MQA-QG's. In the long-document setting, answering a question might need synthesizing many pieces of information from different parts of the document. Therefore, the informativeness property of AttenWalker can be a better method for this setting.

\section{Conclusion}
We study a new task, named unsupervised long-document question answering, and propose AttenWalker, an unsupervised method to incorporate long-range information in QA pairs via graph walking. Extensive experiments show the strong performance of the proposed method. We believe that this work can be an important step in the long-document reasoning with a low-resource setting.

\section*{Limitations}
Despite the strong performance of the proposed AttenWalker. There is still large room for improving efficiency. For example, the time cost of our method is still high. Since we need to search for all Transformer layers and heads to find potentially related spans, the dataset construction could be quite time-consuming. Therefore, an algorithm could be designed in the future to \textit{pre-select} proper layers and heads for attention-based graph walking, which would save much time in dataset construction.

\bibliography{anthology,custom}
\bibliographystyle{acl_natbib}
\clearpage
\appendix

\section*{Appendix}

\section{Analysis of Qasper Question Types}
\label{appendix:analysis_of_qasper_question_types}
In this section, we analyze the contributions of the long document to different question types in the original Qasper dataset. As shown in Table \ref{tab:qasper_question_type_analysis}, when the full text is absent from the input, the performance drops dramatically on the ``Extractive'' and ``Abstractive'' answer types. However, for ``Yes/No'' answers, the performance only drops a little, also keeping a competitive F1 score of 64.84. Besides, the performance of ``Unanswerable'' answers become unexpectedly better. Based on these observations, we argue that ``Yes/No'' and ``Unanswerable'' types are not suitable for testing the ability of long-range reasoning. Therefore, we only use ``Extractive'' and ``Abstractive'' in our experiments. 
\begin{table*}[t]
    \centering
    \begin{tabular}{lcccc|c}    
        \toprule
        Models& Extractive & Abstractive & Yes/No & Unanswerable & Overall\\
        \midrule
        LED +Q +Full Text  & 32.49 & 13.40 & 68.90 & 39.22 & 34.23\\
        LED +Q  & 3.45 & 4.05 & 64.84 & 78.95 & 22.75\\
        \bottomrule   
    \end{tabular}
    \caption{\label{tab:qasper_question_type_analysis} The performance of F1 scores on the dev set of Qasper. In the first row, ``Extractive, Abstractive, Yes/No, Unanswerable'' are four types of answers. ``Overall'' is the F1 score of all the answers. ``LED+Q+Full Text'' denotes training an LED model with a question and the long document as the input. ``LED+Q'' denotes a setting when the question but the long document is not provided for training the QA model.} 
\end{table*}

\section{Details in Fine-Tuning the LED Model}
\label{appendix:details_in_fine_tuning_the_led_model}
Similar to the input setting in \citet{DBLP:conf/naacl/DasigiLBCSG21}, for a long document, we prepend a special token \texttt{</s>} before each paragraph. And then we send the preprocessed long document into an LED model. For example, assume that there is a long document: $\left[t_{1,1}, t_{1, 2}, ..., t_{p, 1}, t_{p,2}, ..., t_{P,P_N-1}, t_{P,P_N}\right]$, where $t_{i, j}$ is the $i$-th token in paragraph $j$, $P$ is the number of paragraphs, $P_N$ is the number of tokens in paragraph $P$. After inserting the special token \texttt{</s>}, the input can be $[\texttt{</s>},t_{1,1}, t_{1, 2}, ..., \texttt{</s>}, t_{p, 1}, t_{p,2}, ..., t_{P,P_N-1}, $ $t_{P,P_N}]$.

\section{Preprocessing Details of Qasper and NarrativeQA}
\label{appendix:preprocessing_details_of_qasper_and_narrativeqa}
\begin{table}[t]
  \begin{center}
      \begin{tabular}{|lccr|}
      \hline
         &  \multirow{2}{*}{\#Examples}  & \multicolumn{2}{c|}{Avg. \#Tokens} \\
         &  & Input & Output \\ \hline \hline
        \multicolumn{4}{|c|}{Qasper} \\ \hline
        Train & 1985 & 5438.6 & 25.8 \\
        Dev & 1393 & 4963.3 & 23.5 \\
        Test & 2695 & 4864.7 & 23.3 \\ \hline \hline
        \multicolumn{4}{|c|}{NarrativeQA} \\ \hline
        Train & 59881 & 74420.1 & 6.0 \\
        Dev & 3461 & 74749.7 & 6.0 \\
        Test & 10557 & 68642.6 & 6.1 \\ \hline
      \end{tabular}
  \end{center}
  \caption{Statistics of Qasper and NarrativeQA.}
  \label{table:datasets_stat}
\end{table}

\subsection{Datasets}
\label{appendix:sec:datasets}
We evaluate the proposed AttenWalker framework on two long-document QA datasets\footnote{The datasets used are originally created for research, which is consistent with our purpose.}: Qasper \citep{DBLP:conf/naacl/DasigiLBCSG21} and NarrativeQA \citep{DBLP:journals/tacl/KociskySBDHMG18}. Qasper \footnote{https://allenai.org/data/qasper} is a dataset (license: CC BY 4.0) for answering questions based on long scientific papers. The questions are annotated based on the abstract of a scientific paper and the answer is annotated by understanding the entire paper's content. The answer types in this dataset can be extractive, abstraction, yes/no or unanswerable. Yet, according to our analysis (Appendix \ref{appendix:analysis_of_qasper_question_types}), QA instances with yes/no or unanswerable answers cannot properly evaluate the ability of long document reasoning. Therefore, we only focus on the extractive and abstractive QA instances in this work. NarrativeQA (license: Apache-2.0) is a QA dataset established upon books and movie scripts of long text sequences. Given summaries of the books/scripts, annotators need to generate corresponding QA pairs where answers are free-formed. Table \ref{table:datasets_stat} shows the statistics of these two datasets.
We use version 0.3 of Qasper dataset\footnote{https://allenai.org/data/qasper} for our experiment, where empty documents are removed. For NarrativeQA, we use the dataset\footnote{https://huggingface.co/datasets/narrativeqa} provided in Huggingface, which is a well-formed dataset. Thus, no extra cleaning step is needed.

\subsection{Unsupervised Long-Document QA Dataset Construction}
\label{appendix:sec:unsupervised_long_document_qa_dataset_construction}

The datasets constructing process is shown in Figure \ref{fig:two_pass_scheme}. Specifically, we first extract sentence constituents from a long document using Berkeley Neural Parser \citep{kitaev-etal-2019-multilingual}. Then, a \texttt{t5-small} model is used in reconstruction-based span selection. In the span linker, we use \texttt{led-base-16384} to acquire the token-level attention graph for span linking. The threshold $\tau$ is set to 0.45. In the answer aggregator, we use the \texttt{bart-large} model to convert spans into an integral answer. Then, an operator \footnote{https://github.com/teacherpeterpan/Unsupervised-Multi-hop-QA} is used to generate questions. In the first pass, the generated dataset is used to train an \texttt{led-base-16384} model. In the second pass, the trained LED model is first used to provide the token-level attention graph as mentioned above. Besides, the global attention scores are also used to complete the attention graph (described in the paragraph ``\textit{Span Graph Constructor}''). The global-attention-related hyperparameters  $K, L, M$. are all set as 3. The construction of the Qasper-document-based dataset costs
12 hours on 4 11GB GPUs while 15 hours on the NarrativeQA-document-based dataset.

\subsection{Long-Document QA Model Setting}
\label{appendix:sec:long_document_qa_model_setting}
We use \texttt{led-base-16384} as the QA model throughout all of our experiments. The input format is described in Appendix \ref{appendix:details_in_fine_tuning_the_led_model}. We searched over batch sizes \{2, 4, 8, 16, 32\}, learning rates \{3e-5, 5e-5, 8e-5, 1e-4\}, warmup proportions \{10\%, 20\%, 30\%, 40\%, 50\%\}, epochs \{2, 4, 5, 6, 8, 10\}. And the final batch size is 16, the learning rate is 5e-5, the warmup proportion is 30\% and the epoch number is 5. We chunk the maximum input length into 13,000 tokens and set the attention window size to 640 so that the LED model in this configuration can be trained on four 11GB GPUs in 3 hours. Despite this relatively limited setting, we find that the performance of the LED model is comparable to the default configuration.

\section{Statistics of the Generated Datasets}
In this section, we summarize the long-document QA datasets generated by AttenWalker. For saving time in QA pair generation, for each document, we randomly sample at most 32 linked span sets for QA-pair generation. The final generated results are shown in Table \ref{tab:stat_dataset}. 
\begin{table}[t]
  \centering
  \begin{tabular}{lrr}
    \toprule
    & Qasper & NarrativeQA\\
    \midrule
    Overall &  22,557 & 25,513\\
    w/ Global Attention &  5,505 & 1,370\\
    Multi-Spans &  10,754 & 8,361\\
    \bottomrule
  \end{tabular}
  \caption{The statistics of QA pairs in the synthetic dataset constructed by AttenWalker.}
  \label{tab:stat_dataset}
\end{table}

\section{Details in the Implementing of Baselines}
\label{appendix:details_in_implementing_baselines}
Since current UQA methods cannot directly apply to the ULQA setting, we make further modifications and describe our implementation in detail.

\paragraph{UNMT \citep{DBLP:conf/acl/LewisDR19}} To generate QA pairs with UNMT, each paragraph in the long document is used as a short context for QA generation. When training the LED model, the question generated by UNMT and the full long document is concatenated into a full sequence so as to train the model.

\paragraph{RefQA \citep{DBLP:conf/acl/LiWDWX20}} Similar to UNMT, each paragraph in the long document is separately used to generate QA pairs.

\paragraph{DiverseQA \citep{DBLP:conf/coling/NieHCM22}} Similar to UNMT and RefQA, each paragraph is selected as a short document. And then, answers of diverse types are extracted from the document. Finally, each question is generated based on the answer and the short document.

\paragraph{MQA-QG \citep{DBLP:conf/naacl/PanCXKW21}} For MQA-QG, in a long document, two paragraphs are randomly sampled. These two paragraphs are then input into the MQA-QG for generating multi-hop QA pairs. Finally, the generated question is concatenated with the long document as the input to train the LED model.

\end{document}